\documentclass[conference]{IEEEtran}
\IEEEoverridecommandlockouts

\usepackage{cite}
\usepackage{amsmath,amssymb,amsfonts}
\usepackage{algorithmic}
\usepackage{graphicx}
\usepackage{textcomp}
\usepackage{xcolor}
\usepackage{booktabs}
\usepackage{multirow}

\def\BibTeX{{\rm B\kern-.05em{\sc i\kern-.025em b}\kern-.08em
    T\kern-.1667em\lower.7ex\hbox{E}\kern-.125emX}}

\begin{document}

\title{Neuron Incidence Redistribution for Fairness in Medical Image Classification}
\author{%
\IEEEauthorblockN{
Abin Shoby, Lyle John Palmer, and Nikhil Cherian Kurian
}
\IEEEauthorblockA{
\textit{Australian Institute for Machine Learning} \\
\textit{Adelaide University} \\
Adelaide, SA 5000, Australia \\
\{abin.shoby, lyle.palmer, nikhil.kurian\}@adelaide.edu.au
}
}
\maketitle

\begin{abstract}
Deep learning models for medical image classification are susceptible to
subgroup performance disparities across demographic attributes such as age,
gender, and race. We identify a latent representational mechanism underlying
these disparities: in transfer-learned models, the dominant penultimate-layer
activation channel under positive predictions is co-activated by both
disease-positive samples and privileged demographic groups (male, older
patients), producing over-diagnosis; conversely, the dominant channel under
negative predictions is co-activated by disadvantaged groups (female, younger
patients), producing systematic under-diagnosis. To address this, we propose
\emph{Neuron Incidence Redistribution} (NIR), a lightweight regularization
method that penalizes the variance of predicted-probability-weighted mean
activations across penultimate-layer neurons, requiring no demographic labels
at training time. On HAM10000, TPR disparity drops from 10.81\% to 0.93\%
across age groups and from 12.04\% to 0.74\% across gender, with a marginal
AUC improvement of 0.51 points. On Harvard OCT-RNFL, NIR reduces FPR
disparity for race (from 15.68\% to 10.66\%) and age (from 12.69\% to
1.80\%), demonstrating that distributing latent disease evidence across the
full penultimate layer is a principled and effective strategy for improving
demographic fairness in medical AI.
\end{abstract}


\section{Introduction}
 
Deep learning classifiers for medical imaging can perform well on average
yet systematically under- or over-diagnose specific demographic groups when
sensitive attributes such as age, gender, or race correlate with disease
prevalence or image morphology \cite{seyyed2020chexclusion}.
Most fairness interventions target output-level disparities in sensitivity
or false-positive rate \cite{zhang2021empirical,zong2023medfair}, but do
not explain how such behavior arises inside the model.
 
We identify a concrete latent mechanism. Inspecting the penultimate-layer
activations of a fine-tuned ResNet-18 reveals an asymmetric pattern of
feature entanglement: the highest-incidence neuron under positive predictions
is jointly co-activated by disease-positive samples \emph{and} privileged
demographic groups (male, older patients), inflating their predicted
probabilities and producing over-diagnosis. Symmetrically, the dominant
channel under negative predictions is preferentially co-activated by
disadvantaged groups (female, younger patients), suppressing their scores
and causing under-diagnosis. This is an instance of \emph{polysemanticity}
\cite{elhage2022toy,goh2021multimodal}---a single feature channel encoding
both clinical and demographic signals simultaneously---and it directly
accounts for the True positive rate (TPR) and False positve rate (FPR) disparities observed at the output level~\cite{seyyed2020chexclusion}.
 
To break this entanglement, we propose \textbf{Neuron Incidence
Redistribution (NIR)}, a regularizer that penalizes the variance of
predicted-probability-weighted mean activations across penultimate-layer
neurons, preventing any single channel from simultaneously dominating
disease prediction and tracking demographic correlates. NIR requires no
demographic labels and adds negligible computational overhead. Our
contributions are: (i)~an empirical analysis revealing that top
positive-class neurons are preferentially co-activated by privileged
demographic groups independent of disease label, linking this
dominant-channel entanglement to asymmetric over- and under-diagnosis;
(ii)~NIR, a differentiable label-free regularizer promoting distributed
positive-class representations across the full penultimate layer; and
(iii)~substantial fairness improvements on HAM10000, with TPR disparity
reductions exceeding 90\% for both age and gender at no AUC cost, and
generally positive fairness improvements on Harvard OCT-RNFL under a
more constrained data regime.
\section{Related Work}

\subsection{Fairness in Medical Imaging}
Demographic disparities have been documented across chest X-ray diagnosis
\cite{seyyed2020chexclusion}, dermatology \cite{groh2021evaluating}, and
ophthalmology \cite{burlina2021addressing}. Mitigation strategies span
pre-processing (dataset rebalancing \cite{chen2021towards}), in-processing
(adversarial fairness \cite{zhang2018mitigating}, fairness-constrained
optimization \cite{lokhande2020fairalm}), and post-processing (threshold
calibration \cite{hardt2016equality}), with MedFair \cite{zong2023medfair}
providing a systematic benchmark across these categories. Unlike these
methods, NIR requires no sensitive attribute supervision, a practical
advantage in clinical deployment.

\subsection{Polysemanticity and Feature Entanglement}
Polysemanticity---individual neurons encoding multiple concepts---is
well-documented in language and vision models
\cite{elhage2022toy,goh2021multimodal}, arising naturally under compression
and sparsity pressures \cite{elhage2022superposition}. In medical imaging,
feature entanglement between disease morphology and demographic attributes
is a plausible pathway to subgroup performance gaps, but has received
limited systematic attention.

\subsection{Representation Regularization}
Latent representation regularization has been explored via disentangled
learning \cite{locatello2019challenging}, information-theoretic constraints
\cite{moyer2018invariant}, and spectral regularization \cite{huang2021fsdr}.
Rather than explicitly disentangling sensitive dimensions, NIR encourages
uniform utilization of the full penultimate layer, reducing the influence
of high-incidence neurons most likely to be polysemantic.
\section{Method}

\begin{figure}
    \centering
    \includegraphics[width=0.8\linewidth]{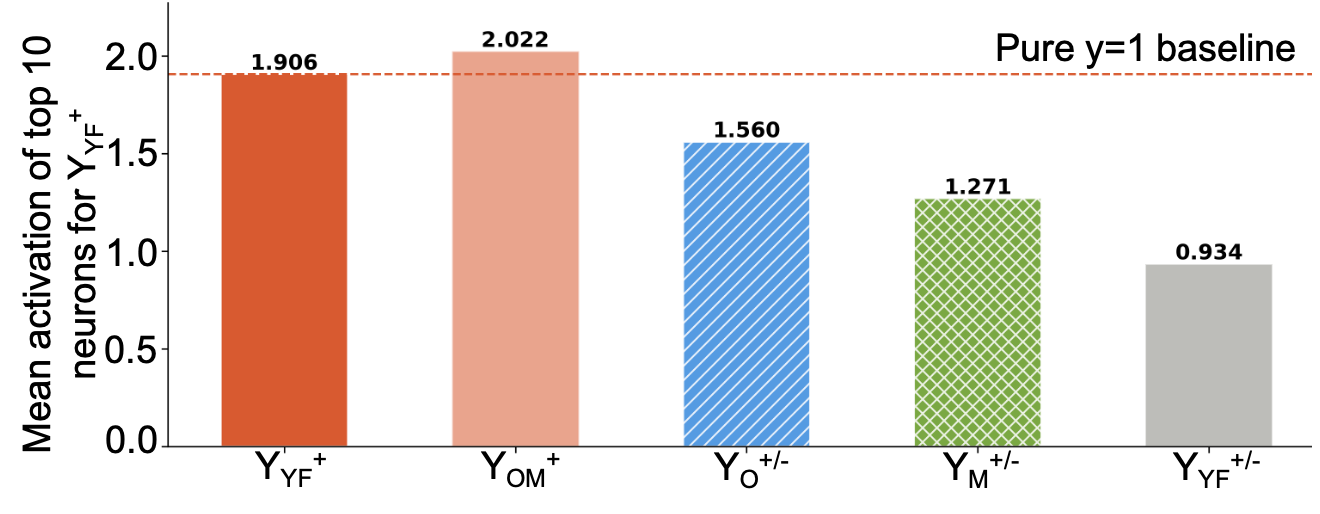}
    \caption{Mean activations of the top 10 positive-class neurons
(selected from disease-positive young female samples, $Y_{YF}^{+}$)
evaluated across demographic subgroups. Disease-positive old male samples
($Y_{OM}^{+}$) produce higher activations in these neurons than the
selecting group itself, indicating that positive-class neurons
polysemantically encode old male demographic features. More broadly,
old ($Y_{O}^{+/-}$) and male ($Y_{M}^{+/-}$) subgroups activate these
neurons more strongly than young female samples ($Y_{YF}^{+/-}$)
regardless of disease label, confirming that demographic correlates are
encoded in superposition with disease morphology.}
    \label{fig:analysis}
\end{figure}

We propose Neuron Incidence Redistribution (NIR), a regularization strategy
designed to reduce concentration of positive-class evidence in a small subset
of penultimate-layer neurons as observed empirically~(Figure~\ref{fig:analysis}).

\subsection{Notation}

Let $\mathcal{D}=\{(\mathbf{x}_i,y_i)\}_{i=1}^{N}$ denote the training
set, where $\mathbf{x}_i \in \mathbb{R}^{H \times W \times C}$ is the
$i$-th input image and $y_i\in\{0,1\}$ is its binary label. For a given
input, the network produces a penultimate-layer representation
$\mathbf{z}_i\in\mathbb{R}^{d}$, a logit $s_i=f(\mathbf{x}_i)$, and a
predicted positive-class probability
$\hat{p}_i=\sigma(s_i)=\frac{1}{1+e^{-s_i}}$.

\subsection{Neuron Incidence}

For a mini-batch of size $B$, let $\mathbf{Z}\in\mathbb{R}^{B\times d}$
denote the matrix of penultimate activations and
$\hat{\mathbf{p}}\in[0,1]^B$ the predicted positive-class probabilities.
We define the \emph{incidence} of neuron $j$ as the
predicted-probability-weighted mean activation:
\begin{equation}
    \phi_j =
    \frac{\sum_{i=1}^{B}\hat{p}_i \, z_{ij}}
         {\sum_{i=1}^{B}\hat{p}_i + \epsilon},
    \qquad j=1,\ldots,d,
    \label{eq:incidence}
\end{equation}
where $\epsilon$ is a small constant for numerical stability. In vector
form,
\begin{equation}
    \boldsymbol{\phi} =
    \frac{\mathbf{Z}^{\top}\hat{\mathbf{p}}}
         {\mathbf{1}^{\top}\hat{\mathbf{p}}+\epsilon}
    \in \mathbb{R}^{d}.
    \label{eq:incidence_vec}
\end{equation}

Weighting by $\hat{p}_i$ ensures that only samples the model already
considers positive contribute to the incidence estimate, focusing the
penalty on the representational structure of predicted positive cases. This
also keeps the loss differentiable and fully compatible with
gradient-based optimization.

\subsection{Incidence Redistribution Loss}

If positive-class evidence is concentrated in only a few neurons, the
variance of $\boldsymbol{\phi}$ will be large. We therefore define the
incidence redistribution loss as the variance of neuron incidences:
\begin{equation}
    \mathcal{L}_{\mathrm{IR}}
    =
    \frac{1}{d}\sum_{j=1}^{d}\left(\phi_j-\bar{\phi}\right)^2,
    \qquad
    \bar{\phi}=\frac{1}{d}\sum_{j=1}^{d}\phi_j.
    \label{eq:lir}
\end{equation}

\subsection{Training Objective}

The final training objective augments the standard Binary Cross-Entropy
(BCE) loss with the NIR regularizer:
\begin{equation}
    \mathcal{L}_{\mathrm{total}}
    =
    \mathcal{L}_{\mathrm{BCE}}
    +
    \lambda \, \mathcal{L}_{\mathrm{IR}},
    \label{eq:total}
\end{equation}
where $\lambda\geq 0$ controls the strength of redistribution. We use a
fixed $\lambda=0.1$ across all experiments to test whether the regularizer
provides consistent fairness improvements without dataset-specific tuning.

\subsection{Interpretation}
 
Minimizing $\mathcal{L}_{\mathrm{IR}}$ discourages \emph{neuron collapse},
wherein a small subset of neurons dominates positive-class representations.
In transfer-learned networks, such dominant neurons tend to be
polysemantic \cite{elhage2022superposition}---encoding disease morphology
and demographic correlates in superposition---producing the asymmetric
over- and under-diagnosis pattern described in Section~I. By distributing
incidence across the full penultimate layer, NIR ensures no single channel
can simultaneously dominate disease prediction and track demographic
attributes, breaking the entanglement at its source without requiring any
demographic supervision.

\section{Experiments}

\subsection{Datasets}

\subsubsection{HAM10000 Skin Lesion Dataset}
We evaluate the proposed method on HAM10000 \cite{tschandl2018ham10000}, a
large public dermoscopic image dataset containing 10,015 images of
pigmented skin lesions collected from multiple dermatoscopy centers. The
dataset includes seven diagnostic categories and has been widely used for
automated skin lesion classification benchmarking. We formulate a binary
classification task distinguishing lesions from normal tissue and evaluate
subgroup disparities across age (binarized at the dataset median into
Young/$\leq$50 and Older/$>$50 groups) and gender. The dataset was split into
training, validation, and test sets using a 70\%/10\%/20\% stratified split
at the image level, preserving the overall class distribution.
Table~\ref{tab:ham_demo} reports the demographic distribution after
preprocessing.

\begin{table}[t]
\centering
\caption{Demographic distribution of HAM10000 after preprocessing.}
\label{tab:ham_demo}
\begin{tabular}{llc}
\toprule
\textbf{Attribute} & \textbf{Group} & \textbf{Count} \\
\midrule
Age & Young ($\leq$50) & 4,153 \\
Age & Older ($>$50)    & 5,795 \\
\midrule
Gender & Male   & 5,400 \\
Gender & Female & 4,548 \\
\bottomrule
\end{tabular}
\end{table}

\subsubsection{Harvard OCT-RNFL Glaucoma Dataset}
We further evaluate on the Harvard Glaucoma Detection and Progression
dataset \cite{luo2024harvard}, which provides retinal nerve fiber layer
thickness (RNFLT) OCT imaging for glaucoma detection. The binary
classification task distinguishes glaucoma from non-glaucoma cases. This
dataset is well-suited for fairness evaluation because it includes
demographic annotations (sex and race) and because prior work has
highlighted racial disparities in glaucoma screening \cite{burlina2021addressing}.
We evaluate subgroup disparities across race (White, Black, Asian) and
gender, using 70\%/10\%/20\% stratified split protocol.
Table~\ref{tab:rnflt_demo} reports the demographic distribution.

\begin{table}[t]
\centering
\caption{Demographic distribution of the Harvard OCT-RNFL dataset after
preprocessing.}
\label{tab:rnflt_demo}
\begin{tabular}{llc}
\toprule
\textbf{Attribute} & \textbf{Group} & \textbf{Count} \\
\midrule
Age & Young ($\leq$61) & 1650 \\
Age & Older ($>$61) & 1650 \\
\midrule
Race & White & 1100 \\
Race & Black & 1100 \\
Race & Asian & 1100 \\
\midrule
Gender & Male   & 1650 \\
Gender & Female & 1812 \\
\bottomrule
\end{tabular}
\end{table}

\subsection{Model Architecture and Training}
Both models use an ImageNet-pretrained ResNet-18 \cite{he2016deep} with
the final layer replaced by a single-output linear classifier with sigmoid
activation, fine-tuned end-to-end using the Adam optimizer
\cite{kingma2015adam} at a learning rate of $3\times10^{-3}$ for 30 epochs
with early stopping on validation AUC. The baseline uses standard BCE loss;
NIR augments this with the objective in \eqref{eq:total} at $\lambda=0.1$.
All other hyperparameters, augmentation, and random seeds are held constant
to isolate the effect of the regularizer.

\subsection{Evaluation Metrics}

We assess both classification performance and subgroup fairness.
Classification performance is measured by the area under the receiver
operating characteristic curve (AUC). The decision threshold is selected on the validation set by maximizing
Youden's J statistic ($J = \mathrm{TPR} - \mathrm{FPR}$), defined as the
point on the ROC curve that maximises the sum of sensitivity and specificity,
and is then held fixed for test evaluation~\cite{seyyed2020chexclusion}.

Subgroup fairness is measured by the absolute disparity in true-positive
rate (TPR) and false-positive rate (FPR) across demographic subgroups:
\begin{equation}
    \Delta_{\mathrm{TPR}}
    =
    \max_g \mathrm{TPR}_g - \min_g \mathrm{TPR}_g,
    \label{eq:dtpr}
\end{equation}
\begin{equation}
    \Delta_{\mathrm{FPR}}
    =
    \max_g \mathrm{FPR}_g - \min_g \mathrm{FPR}_g.
    \label{eq:dfpr}
\end{equation}

$\Delta_{\mathrm{TPR}}$ captures potential under-diagnosis disparities;
$\Delta_{\mathrm{FPR}}$ captures potential over-diagnosis disparities.
Lower values indicate fairer subgroup performance.

\section{Results and Discussion}

\subsection{Subgroup Fairness}

Table~\ref{tab:disparity_results} reports TPR and FPR disparity for the
baseline (BCE) and NIR-regularized model across both datasets and all
demographic subgroups. NIR yields substantial fairness improvements on
HAM10000 for both age and gender. TPR disparity across age groups decreases
from 10.81\% to 0.93\%, and across gender from 12.04\% to 0.74\%---reductions of
91.4\% and 93.8\%, respectively. FPR disparity also improves in both cases.

On the Harvard OCT-RNFL dataset, results are more mixed. NIR substantially reduces FPR disparity for age (from 12.69\% to 1.80\%) and race (from 15.68\% to 10.66\%). TPR disparities improve modestly for gender (from 7.23\% to 5.61\%) and race (from 3.36\% to 3.03\%), and partially for age (from 11.76\% to 9.42\%). Gender FPR disparity increases from 2.23\% to 4.79\%, reflecting the limits of soft redistribution under small-sample and imbalanced conditions. The relatively small dataset size and racial group imbalance likely constrain NIR's ability to fully redistribute incidence uniformly across all subgroups.

\begin{table}[t]
\centering
\caption{TPR and FPR disparity across datasets, subgroups, and methods.
$\downarrow$ indicates lower is better. Best results per row in
\textbf{bold}.}
\label{tab:disparity_results}
\renewcommand{\arraystretch}{1.3}
\begin{tabular}{llccc}
\toprule
\textbf{Dataset} & \textbf{Subgroup} & \textbf{Method}
    & \textbf{$\Delta_{\mathrm{TPR}}$\,(\%)\,$\downarrow$}
    & \textbf{$\Delta_{\mathrm{FPR}}$\,(\%)\,$\downarrow$} \\
\midrule
\multirow{4}{*}{HAM10000}
    & \multirow{2}{*}{Age}
        & BCE  & 10.81 & 12.82 \\
    &   & NIR  & \textbf{0.93}  & \textbf{12.32} \\
\cmidrule{2-5}
    & \multirow{2}{*}{Gender}
        & BCE  & 12.04 & 7.16 \\
    &   & NIR  & \textbf{0.74}  & \textbf{4.12} \\
\midrule
\multirow{6}{*}{RNFLT}
    & \multirow{2}{*}{Age}
        & BCE  & 11.76 & 12.69 \\
    &   & NIR  & \textbf{9.42}  & \textbf{1.80} \\
\cmidrule{2-5}
    & \multirow{2}{*}{Gender}
        & BCE  & 7.23  & \textbf{2.23} \\
    &   & NIR  & \textbf{5.61}  & 4.79 \\
\cmidrule{2-5}
    & \multirow{2}{*}{Race}
        & BCE  & 3.36  & 15.68 \\
    &   & NIR  & \textbf{3.03}  & \textbf{10.66} \\
\bottomrule
\end{tabular}
\end{table}

\subsection{Classification Performance}

Table~\ref{tab:auc_comparison} reports AUC for both models. On HAM10000,
NIR marginally improves AUC (89.74$\to$90.25), suggesting that a more
distributed latent representation does not harm discriminative performance
and may even provide mild regularization benefits.

On the RNFLT dataset, NIR incurs a 3.03-point AUC reduction (84.71$\to$81.68).
We attribute this to the smaller dataset size and greater inter-group
imbalance of the RNFLT cohort, where enforcing uniform neuron incidence may
partially suppress high-incidence neurons that carry genuine disease
information alongside demographic correlates. This reflects an inherent
accuracy--fairness tradeoff: forcing broader feature utilization can reduce
reliance on the most discriminative (but demographically entangled) neurons.
We regard this tradeoff as an important consideration for future work and
discuss it in Section~\ref{sec:discussion}. Notably, despite the AUC reduction, NIR still achieves meaningful FPR disparity improvements on RNFLT, indicating partial fairness benefit even under a constrained data regime

\begin{table}[t]
\centering
\caption{AUC comparison. $\uparrow$ indicates higher is better.}
\label{tab:auc_comparison}
\begin{tabular}{lccc}
\toprule
\textbf{Dataset}
    & \textbf{BCE (AUC\,$\uparrow$)}
    & \textbf{NIR (AUC\,$\uparrow$)}
    & \textbf{$\Delta$AUC} \\
\midrule
HAM10000 & 89.74 & \textbf{90.25} & +0.51 \\
RNFLT    & \textbf{84.71} & 81.68 & $-$3.03 \\
\bottomrule
\end{tabular}
\end{table}

\section{Discussion}
\label{sec:discussion}
 
\paragraph{Mechanistic account.}
As established in Section~I, the baseline model's dominant positive-class
channel is co-activated by disease labels and privileged demographics
(male, older), inflating their scores and causing over-diagnosis; the
dominant negative-class channel co-activates disadvantaged groups (female,
younger), suppressing their scores and causing under-diagnosis. NIR's
variance penalty prevents any single channel from concentrating the
positive-class signal, decoupling classifier sensitivity from subgroup
membership without requiring demographic supervision.
 
\paragraph{Limitations.}
The AUC drop on RNFLT reflects an accuracy--fairness tradeoff under limited
data: NIR may suppress high-incidence neurons that carry genuine disease
signal alongside demographic correlates. NIR is also a soft incentive and
cannot guarantee redistributed features are demographically independent;
future work should pair it with explicit disentanglement or causal
constraints. Extension to multi-class settings requires generalizing the
incidence formulation beyond binary predictions.
\section{Conclusion}

We presented Neuron Incidence Redistribution (NIR), a simple regularization
method that improves demographic fairness in medical image classifiers by
discouraging concentration of positive-class evidence in sparse, potentially
polysemantic penultimate-layer neurons. NIR requires no demographic labels
during training and adds negligible computational overhead. On HAM10000, it
reduces TPR disparity by over 90\% for both age and gender while marginally
improving AUC. Results on the smaller RNFLT dataset reveal an
accuracy--fairness tradeoff that warrants further investigation. We believe
that connecting representation-level mechanisms---such as polysemanticity
and neuron collapse---to observable demographic disparities is a promising
direction for principled and interpretable fairness research in medical AI.


\section*{Acknowledgment}
We thank GSK AI for supporting this work through the Responsible AI Research Grant, and the Australian Institute for Machine Learning (AIML), Adelaide University, for institutional support and research infrastructure.


\end{document}